\pdfoutput=1

\documentclass[11pt]{article}
\usepackage[dvipdfmx]{graphicx}

\usepackage{EACL2023}

\usepackage{times}
\usepackage{latexsym}

\usepackage{amsfonts}
\usepackage{amsmath}
\usepackage{bm}
\usepackage{booktabs}
\usepackage{textcomp}
\usepackage{wrapfig}

\usepackage[T1]{fontenc}

\usepackage[utf8]{inputenc}

\usepackage{microtype}

\usepackage{inconsolata}

\title{Learning by Asking Questions for Knowledge-based Novel Object Recognition}

\author{Kohei Uehara \\
  The University of Tokyo \\
  \texttt{uehara@mi.t.u-tokyo.ac.jp} \\\And
  Tatsuya Harada \\
  The University of Tokyo \\
  RIKEN \\
  \texttt{harada@mi.t.u-tokyo.ac.jp} \\}

\begin{document}
\maketitle

\begin{abstract}
In real-world object recognition, there are numerous object classes to be recognized.
Conventional image recognition based on supervised learning can only recognize object classes that exist in the training data, and thus has limited applicability in the real world.
On the other hand, humans can recognize novel objects by asking questions and acquiring knowledge about them.
Inspired by this, we study a framework for acquiring external knowledge through question generation that would help the model instantly recognize novel objects.
Our pipeline consists of two components: the Object Classifier, which performs knowledge-based object recognition, and the Question Generator, which generates knowledge-aware questions to acquire novel knowledge.
We also propose a question generation strategy based on the confidence of the knowledge-aware prediction of the Object Classifier.
To train the Question Generator, we construct a dataset that contains knowledge-aware questions about objects in the images.
Our experiments show that the proposed pipeline effectively acquires knowledge about novel objects compared to several baselines.
\end{abstract}

\section{Introduction}
\label{sec:intro}
Object category recognition has long been a central topic in computer vision research.
Traditionally, object recognition has been addressed by supervised learning using a large dataset of image-label pairs~\cite{Deng2009ImageNetAL}.
However, with supervised approaches, the model can only recognize a frozen set of object classes and is not suitable for real-world object recognition, where numerous object classes exist.
Recently, image recognition methods based on contrastive learning using image-text pair datasets have emerged~\cite{clip, align}.
By training on hundreds of millions of image-text pairs, these models have acquired remarkable zero-shot recognition capabilities for a wide variety of objects.
However, these models can recognize objects that commonly appear in the pre-training dataset but are not as effective for rare objects~\cite{Shen2022KLITELT}.
Collecting new data and retraining the entire model to make these models recognize novel objects is impractical considering the cost of data collection and computation.
Therefore, it is essential to develop a method that enables the model to recognize novel objects while maintaining low data collection costs and avoiding model retraining as much as possible.

\begin{figure*}[t]
   \centering
   \includegraphics[width=1.0\linewidth]{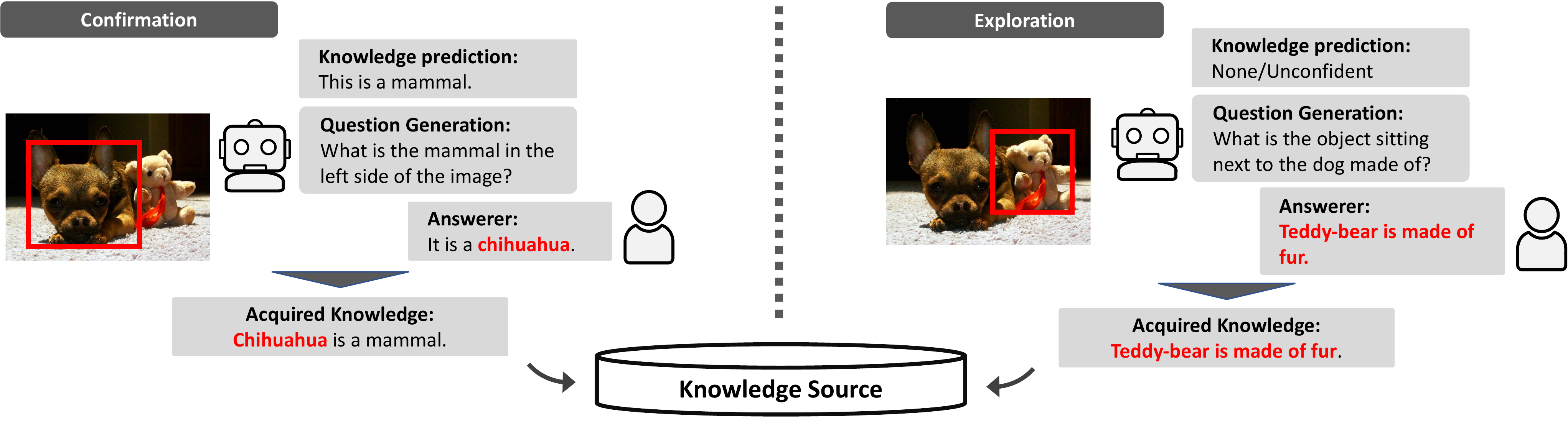}
   \caption{
       Conceptual illustration of our proposed pipeline.
   If the model is confident about the predicted knowledge, question generation is performed in confirmation mode.
If the model is not confident, question generation is performed in exploration mode.
   }
   \label{fig:intro}
\end{figure*}

When humans acquire knowledge about the world, asking questions and explicitly acquiring knowledge are important skills involved~\cite{chouinard2007children,Ronfard2018QuestionaskingIC}.
Inspired by this, we explored methods to dynamically increase knowledge in image recognition by asking questions.
This approach has several advantages over the traditional supervised learning method: (1) it requires only a small amount of data to acquire knowledge because the system acquires only the knowledge it needs, and (2) it has a low data collection cost because the system itself seeks the required data.

We propose a pipeline consisting of a knowledge-based object classifier (OC) and a question generator (QG) for knowledge acquisition.
Following previous research on structured knowledge~\cite{Ji2022ASO}, we represent knowledge as a knowledge triplet, that is, a list of three words or phrases: \textit{head}, \textit{relation}, and \textit{tail}, such as $\langle$dog, IsA, mammal$\rangle$.
We train the OC to retrieve knowledge from knowledge sources, which outputs the corresponding head in the knowledge source as the predicted object class (e.g., \textlangle IsA, mammal\textrangle$\,\to\,$dog).
The QG model then generates questions to add new knowledge to the knowledge source for novel object recognition.
In the QG model, we use two modes in question generation: \textbf{confirmation} and \textbf{exploration}, as illustrated in Figure~\ref{fig:intro}.
First, ``confirmation'' is used when the unknown object is relatively close to a known object category.
For example, if the model knows about ``dog,'' then a novel category ``chihuahua'' is considered to be a close concept to ``dog.''
In this case, the model can infer reasonable knowledge (e.g., both ``chihuahua'' and ``dog'' are a type of mammal) and ask questions to confirm it,  such as ``What is the mammal on the left side of the image?''
In contrast, the ``exploration'' mode is used when the unknown object is far from the existing object category (e.g., ``teddy-bear'' may not resemble any known object class).
In this case, the model is unable to estimate the proper knowledge and attempts to obtain all the necessary knowledge by asking questions (``What is the object sitting next to the dog made of?'').

Our contributions and findings can be summarized as follows:
\begin{itemize}
    \setlength{\parskip}{-2pt}
    \setlength{\itemsep}{3pt}
    \item We propose a novel pipeline to acquire knowledge about novel objects by asking questions. We designed the OC model based on CLIP~\cite{clip} and the QG model as a Transformer~\cite{transformer} based text generation model.
    \item We built a novel dataset to train the QG model, namely, \textbf{Professional K-VQG}. This dataset contains a variety of annotations such as object labels, bounding boxes, knowledge, and knowledge-aware questions.
    \item We compare our proposed pipeline with several baselines and show that the knowledge acquired through question generation is effective for novel object recognition.
\end{itemize}

\section{Related Work}\label{sec:related-works}
\paragraph{Novel object recognition}
Increasing the number of recognizable object classes is a widely studied problem in object recognition.
A typical approach in novel object recognition is to train a model that computes the similarity between the visual and semantic features of objects.
To compute semantic features of a novel object, external knowledge about the object (e.g., attributes~\cite{Lampert2009LearningTD,Farhadi2009DescribingOB, Jayaraman2014ZeroshotRW, Akata_2016_CVPR,Li2021ZeroshotRW}, class hierarchy~\cite{Rohrbach2011EvaluatingKT, Wang2018ZeroShotRV}, or textual description~\cite{Ba2015PredictingDZ,Qiao2016LessIM,Reed2016LearningDR,Zareian2021OpenVocabularyOD}) is often used.
Recently proposed vision-and-language contrastive learning methods, such as CLIP~\cite{clip} or ALIGN~\cite{align}, use extremely large-scale image caption data to learn the relationship between images and their textual descriptions.
With the help of the prefix-tuning technique, these models exhibited a strong zero-shot recognition ability.
However, the abovementioned studies have a problem in that they require either a well-prepared knowledge database on novel objects or a large number of image-text pair datasets and appropriately designed prompts, both of which are labor-intensive tasks for humans.
In our method, once the question generation model is trained, the model dynamically acquires the necessary knowledge, thereby reducing human effort.

\paragraph{Learning by asking (LBA)}
LBA generates questions to collect additional data to train a model.
With the development of natural language generation methods, several studies using question generation to acquire the information necessary to solve a task (e.g., reading comprehension~\cite{Du2017LearningTA,Yuan2017MachineCB} or question answering~\cite{scialom-staiano-2020-ask}) have been conducted.
In addition, in vision-and-language fields, LBA is applied to VQA tasks~\cite{lba} or image captioning tasks~\cite{vqg_caption}.
However, our research is the first attempt to apply LBA to a novel object recognition task, which is a more real-world-oriented visual recognition task.

\paragraph{Visual question generation (VQG)}
The first studies to address VQG used a simple method of inputting image features into a text decoder and generating questions~\cite{vqg}.
Subsequent research has focused on how to better control the content of the questions to be generated.
In general, the control of question generation is achieved by providing additional information to the text decoder in addition to the image features.
The main methods include providing the answers~\cite{iqan,ivqa} and providing categories of the answers~\cite{info-vqg,c3vqg,vqg_unknown}.
In addition, similar to this study, another study targeted knowledge that can be acquired through questioning~\cite{Uehara2022KVQGKV}.
They created a knowledge-aware VQG dataset (\textbf{K-VQG}) using AMT and used UNITER~\cite{uniter}, which is a state-of-the-art vision-and-language transformer, as an encoder for images and knowledge to successfully generate questions for the knowledge.
We designed a question generation model based on their work and created a new dataset as an extension of their dataset.

\section{Method}\label{sec:method}

\begin{figure*}[t]
    \centering
    \includegraphics[width=0.7\linewidth]{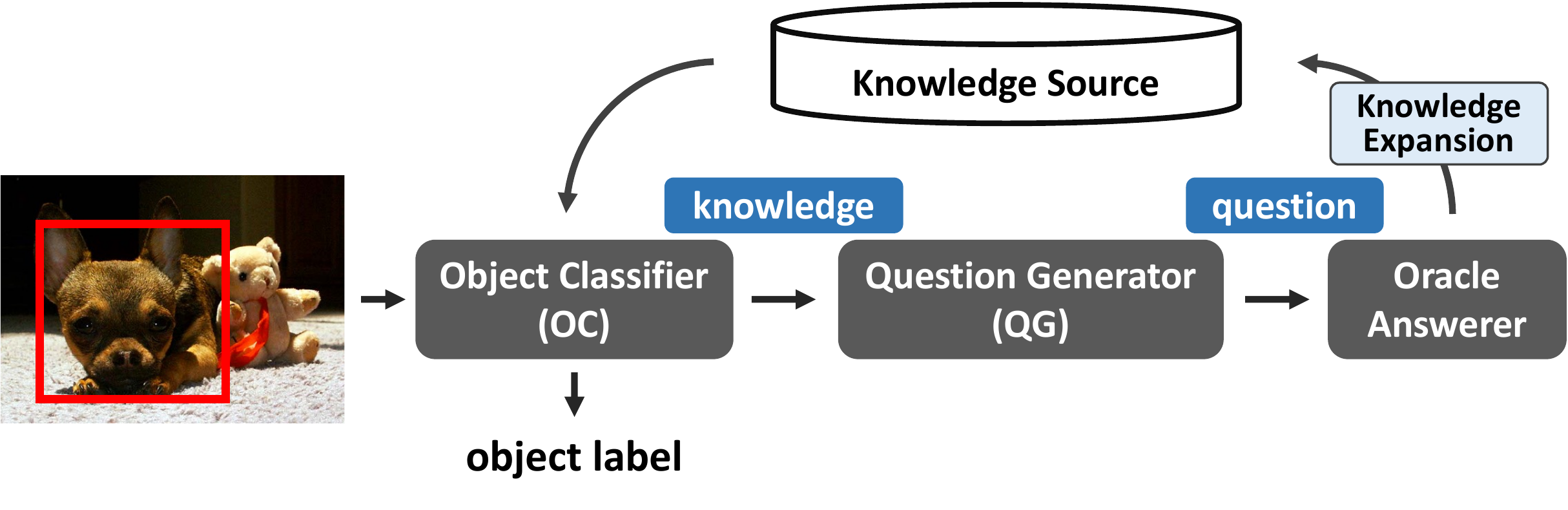}
   \caption{
       Overall pipeline of our method.
   The OC model performs knowledge-based object recognition using knowledge sources.
   The QG model generates questions target the knowledge needed for novel object recognition.
   Answers to the questions are provided by the Oracle Answerer and added to the knowledge source.
   With the newly added knowledge, the OC model is able to recognize novel objects.
   }
   \label{fig:overall}
\end{figure*}

Our system consists of an \textbf{O}bject \textbf{C}lassifier (OC) and a \textbf{Q}uestion \textbf{G}enerator (QG).
First, we describe the overall pipeline of our system (Figure~\ref{fig:overall}).
Then, we describe the detail of each module in the following sections.

\begin{figure}[t]
    \centering
    \includegraphics[width=1.0\linewidth]{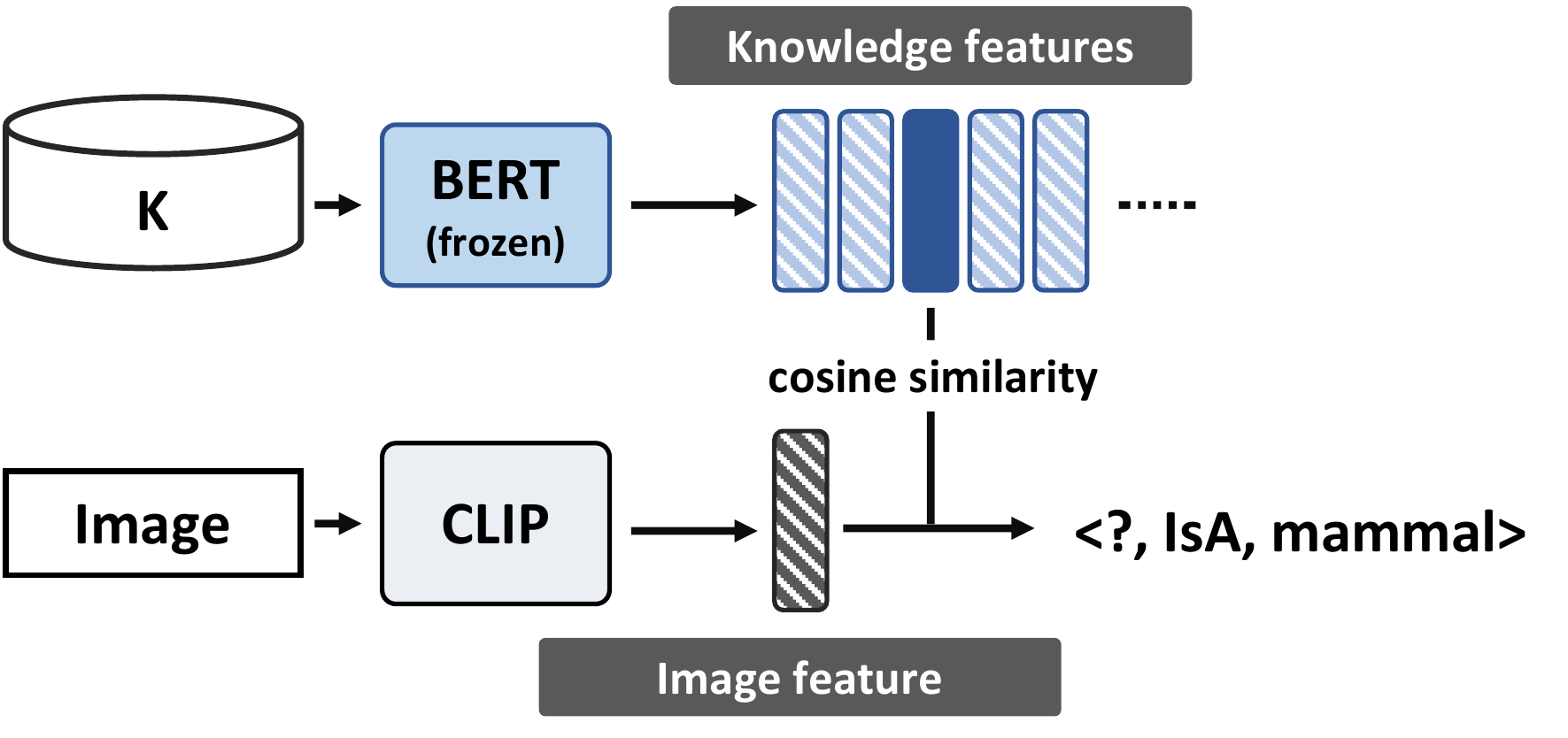}
   \caption{
       Architecture of the OC model.
   Based on the knowledge encoded by BERT and the similarity calculation of the object image encoded by CLIP, the prediction of the knowledge required for object recognition is performed.
   }
   \label{fig:classifier}
\end{figure}

The OC model explicitly utilizes knowledge when object recognition is performed.
In other words, for a given image of the object, the OC retrieves the appropriate knowledge $k = [h,\;r,\;t] \in \mathcal{K}$, where $h$, $r$, and $t$ are the head, relation, and tail, respectively, and $\mathcal{K}$ is a knowledge source.
Here, $h$ corresponds to the object label.
Thus, once the model predicts the knowledge of the object image, it can search $\mathcal{K}$ using the predicted knowledge and obtain the object label.
This design allows the model to instantly recognize novel objects when it obtains knowledge about them, simply by updating the knowledge source, without retraining the model.

The QG model is responsible for generating questions about the objects in the image and acquiring knowledge that is useful for novel object recognition.
To this end, we conditioned the QG with partial knowledge, which masks part of the knowledge.

Once the answers to the generated questions are obtained, the acquired knowledge $\mathcal{K}'$ is added to the model's original knowledge source $\mathcal{K}$.
Consequently, the OC's knowledge source is updated to $\mathcal{K}^+ = \mathcal{K}\;\cup\;\mathcal{K}'$.
Then, in the next inference phase, the OC makes predictions about the knowledge and labels by referring to the updated knowledge source $\mathcal{K}^+$.

\subsection{Object Classifier}\label{subsec:object-classifier}

The OC model (Figure~\ref{fig:classifier}) predicts object-related knowledge by computing the similarity between the object feature $\bm{f}_o \in \mathbb{R}^{d}$ and the knowledge feature $\bm{f}_{k} \in \mathbb{R}^{d}$ of the associated knowledge, that is, $p(k) = \mathrm{sim} (\bm{f}_o,\;\bm{f}_k)$, where $d$ is the dimension of the object features and knowledge features.

We adopted a model based on CLIP~\cite{clip}, which is one of the latest visual recognition models, as the object knowledge predictor.
CLIP is a model consisting of an image encoder and text encoder, each of which can be used to compute the similarity between images and text.
The image encoder of CLIP $f_{\theta}$ receives a cropped image $I_{\textrm{crop}}$ as input and outputs the visual feature $\bm{f}_o$.
Knowledge features of knowledge $\bm{f}_k$ is also computed using pre-trained CLIP text encoder $f_{\phi}$.
Here, we converted the knowledge from a triplet representation (e.g., <cat, IsA, mammal>) to the representation of a single sentence with a masked head (e.g., ``[MASK] is a mammal'') before feeding it into the text encoder.
We used cosine similarity to compute the similarity between the object and knowledge features as follows:
\begin{align}
    \bm{f}_o = f_{\theta}(I_{crop}),\quad
    \bm{f}_k = f_{\phi}(k) \\
    \mathrm{sim}(\bm{f}_o,\;\bm{f}_k) = \frac{\bm{f}_o^{\top} \bm{f}_k}{\|\bm{f}_o\| \|\bm{f}_k\|}
\end{align}

The OC model is trained to minimize the binary cross-entropy loss as follows:
\begin{multline}
L_{\mathrm{OC}} =  -\sum^{|\mathcal{K}|}_i \bigl(
 y_i \cdot \log \sigma (\mathrm{sim}(\bm{f}_o,\;\bm{f}_{k_i})) + \\ (1-y_i) \cdot \log (1-\sigma (\mathrm{sim}(\bm{f}_o,\;\bm{f}_{k_i}))) \bigr)
\end{multline}
where $y_i \in \{0, 1\}$ indicates the ground-truth label for the $i$-th knowledge.

If the model successfully predicts the knowledge, then it has the relation and tail of the knowledge about the object.
Then, when inferring labels from the predicted knowledge $\hat{k}$, we search the knowledge source $\mathcal{K}$ for knowledge that satisfies the predicted relation and tail conditions and treat the matching head as the predicted label.

\subsection{Question Generator}\label{subsec:question-generator}
The QG model takes an image $I$ and target knowledge $k$ as input and generates question $q$.

First, we explain how the target knowledge is determined.
In knowledge acquisition, it is important to acquire knowledge that is ``appropriate'' and ``useful'' for recognition, that is, to acquire correct knowledge at the lowest possible cost.
Here, ``low cost'' implies that retraining the OC model should be avoided as much as possible.
Therefore, we propose using two different modes of question generation: ``confirmation'' and ``exploration.''
As described in Sec.~\ref{sec:intro}, the ``confirmation'' mode is used when the unknown object is relatively close to a known object category, whereas the ``exploration'' mode is used when the unknown object is far from the existing object category.
The target knowledge $k$ in each case is defined as follows:

\begin{equation}
    k = \left\{
    \begin{array}{ll}
        \lbrack\,\texttt{MASK},\;\hat{r},\;\hat{t}\,\rbrack & \mathrm{(confirmation)}\\
        \lbrack\,\texttt{MASK},\;r^*,\;\texttt{MASK}\,\rbrack & \mathrm{(exploration)}
    \end{array}
    \right.\label{eq:knowledge}
\end{equation}

Here, $\hat{r}$ and $\hat{t}$ are the predicted relation and tail, respectively, and $r^*$ is an arbitrarily chosen relation based on the frequency of the relation in the data.

We control the policy of mode selection based on the expected value of the utility that the model can obtain from the answer.
Here, we define the policy selection function $\pi$, which takes the value 1 for the confirmation mode and 0 for the exploration mode.
Then, we adopt a policy that maximizes the expected utility of the model with utility function $u_{\theta}$ for the training data $\mathcal{X}$:
\begin{equation}
    \frac{1}{|\mathcal{X}|}\sum (\pi \, u_{\theta} (\bm{f}_o) + (1 - \pi) \, u_{\theta} (\bm{f}_o))\label{eq:expected_utility}
\end{equation}

We define the utility function as the sum of the ``correctness'' and ``informativeness'' of the expected answer.
The ``correctness'' represents the estimated correctness of the knowledge expected to be acquired by the answer.
For simplicity, we assume that the oracle answer should be correct and suppose that the expected correctness is 1.0 when the mode is ``exploration.''
On the other hand, when the mode is ``confirmation,'' the expected correctness depends on the confidence of the model $\mathrm{conf}(\hat{k})$; thus, we set the expected correctness as the predicted score output by the OC model.

The ``informativeness'' is the value representing how useful the acquired knowledge is to the model.
For the ``exploration'' mode, we estimate the informativeness using the similarity between the input image feature and target knowledge feature $\mathrm{sim} (\bm{f}_o,\;\bm{f}_{\hat{k}})$.
For the ``confirmation'' mode, we use the expected value of the similarity based on the mean similarity of the training data, i.e., $\mathbf{E}[I] \simeq \frac{1}{|\mathcal{X}|}\sum \mathrm{sim} (\bm{f}_{o},\;\bm{f}_{\hat{k})}$.

The utility function is expressed as follows:
\begin{equation}
    u_{\theta}(\bm{f}_o) = \left\{
    \begin{array}{ll}
        \mathrm{conf}(\hat{k}) + \mathrm{sim} (\bm{f}_o,\;\bm{f}_{\hat{k}}) & \mathrm{(conf.)}\\
        1 + \frac{1}{|\mathcal{X}|} \sum \mathrm{sim} (\bm{f}_{o},\;\bm{f}_{\hat{k}}) & \mathrm{(exp.)}
    \end{array}
    \right.\label{eq:utility}
\end{equation}

Once the input knowledge $k$ is determined, question generation is performed using it as input.
In the question generation model, we used the ViLT~\cite{vilt}  encoder, which is one of the state-of-the-art vision-and-language model, and GPT-2~\cite{gpt2} based decoder.
The ViLT encoder $\mathrm{Enc}(\cdot)$ takes two inputs: (1) the input image $I$ and the masked region image $I_M$ and (2) knowledge triplets in sentence form, i.e., ``[MASK] is a mammal.''
Here, a masked region image is one in which all pixel values outside the target region are replaced by zero.
A learnable type embedding vector is added to the masked image and the original image, respectively.

The model is trained using the following loss function:
\begin{equation}
    L = - \sum_{i} \log P(y_i \; | \; y_{<i},\; \mathrm{Enc}([I, I_M],\;k))\label{eq:vqg_loss}
\end{equation}

Here, $y_i$ is the $i$-th word of the question and $\mathrm{Enc}(\cdot)$ is the ViLT encoder, which produces fused feature of the visual and textual knowledge features.

\subsection{Oracle Answerer}\label{subsec:oracle-answerer}

\begin{figure}[t]
    \centering
    \includegraphics[width=0.7\linewidth]{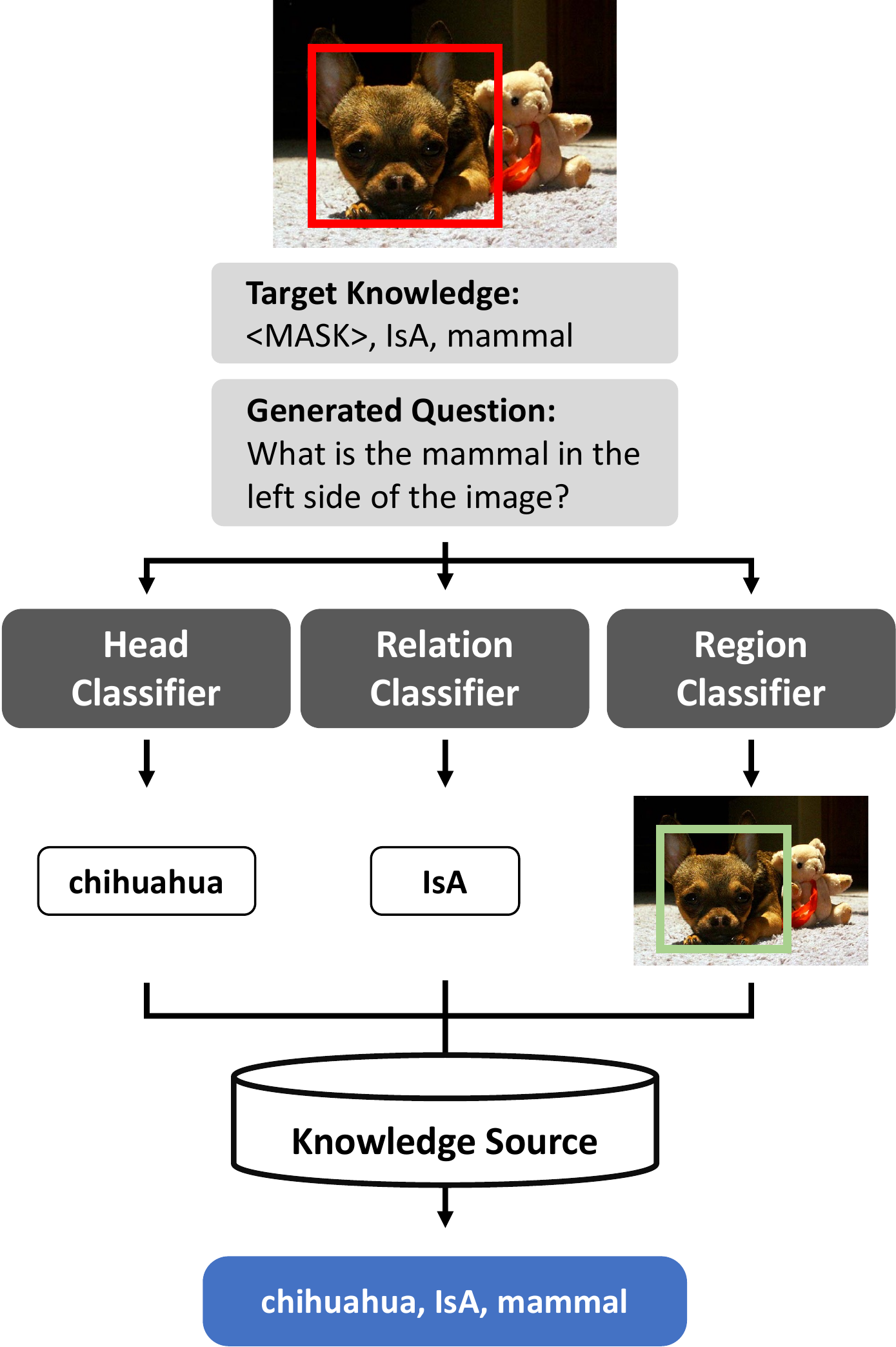}
    \caption{The architecture of the oracle answerer.}
\label{fig:oracle_answerer}
\end{figure}

Given an image and a generated question, the Oracle Answerer predicts the answer knowledge to the question.
We implement this module as a composition of three sub-modules; (1) Head classifier, (2) Relation classifier, and (3) Region classifier.
Each module checks whether the generated question is ``valid'', and if all modules agree that the question is ``valid,'' Oracle Answerer searches the oracle knowledge source and outputs the knowledge that matches the targeted head and relation.
Oracle knowledge source is a knowledge source that merges ConceptNet~\cite{conceptnet} and \textsc{Atomic$^{20}_{20}$}~\cite{comet2020}, which were used in the creation of the datset (see Sec.~\ref{subsec:dataset}).

\paragraph{Head classifier}
The head classifier $\mathcal{H}$ predicts the head of the target knowledge from the generated question, i.e., $h = \mathcal{H}(I, Q)$.
We implement this module following the standard VQA methodology, that is, as a multi-class classification problem that outputs the proper entity given an image and a question.
For this module, we fine-tuned pre-trained ViLT-VQA~\cite{vilt} model.
This module returns ``valid'' if the predicted head is equal to the object in the target region.

\paragraph{Relation classifier}
The relation classifier $\mathcal{R}$ predicts the relation of the target knowledge from the generated question, i.e., $r = \mathcal{R}(Q)$.
Since this problem can be formulated as a sentence classification problem, we use a fine-tuned Distil-BERT~\cite{distilbert} as the relation classifier.
This module returns ``valid'' if the predicted relation matches the target relation ($r$ in the eq. \ref{eq:knowledge}).

\paragraph{Region classifier}
The region classifier $\mathcal{G}$ predicts the target region, i.e., $g = \mathcal{G}(I, Q)$.
We design this module as a model that, given a question and a set of candidate regions, outputs the region that is most relevant to the question.
This problem setup is similar to that of the Referring Expression Comprehension (RE Comprehension)~\cite{refcoco}.
Therefore, we used a fine-tuned version of UNITER grounding model~\cite{uniter}, which has achieved high performance in RE Comprehension task.
This module returns ``valid'' if the predicted region is sufficiently close to the target region.
We calculated the IoBB (Intersection over Bounding Box) between the predicted region and the target region and considered two regions sufficiently close if the value was greater than 0.4.

\subsection{Knowledge Expansion}\label{subsec:knowledge-expansion}
When an answer knowledge $k`$ is obtained for a generated question $q$ by the model, it is added to the model's knowledge source $\mathcal{K}$, that is, $\mathcal{K}^+ = \mathcal{K}\;\cup\;\{k'\}_{i=1}^M$, where $M$ is the number of newly acquired knowledge.

\section{Experiments}\label{sec:experiments}

\subsection{Dataset}\label{subsec:dataset}

\begin{figure}[t]
\centering
\includegraphics[width=1.0\linewidth]{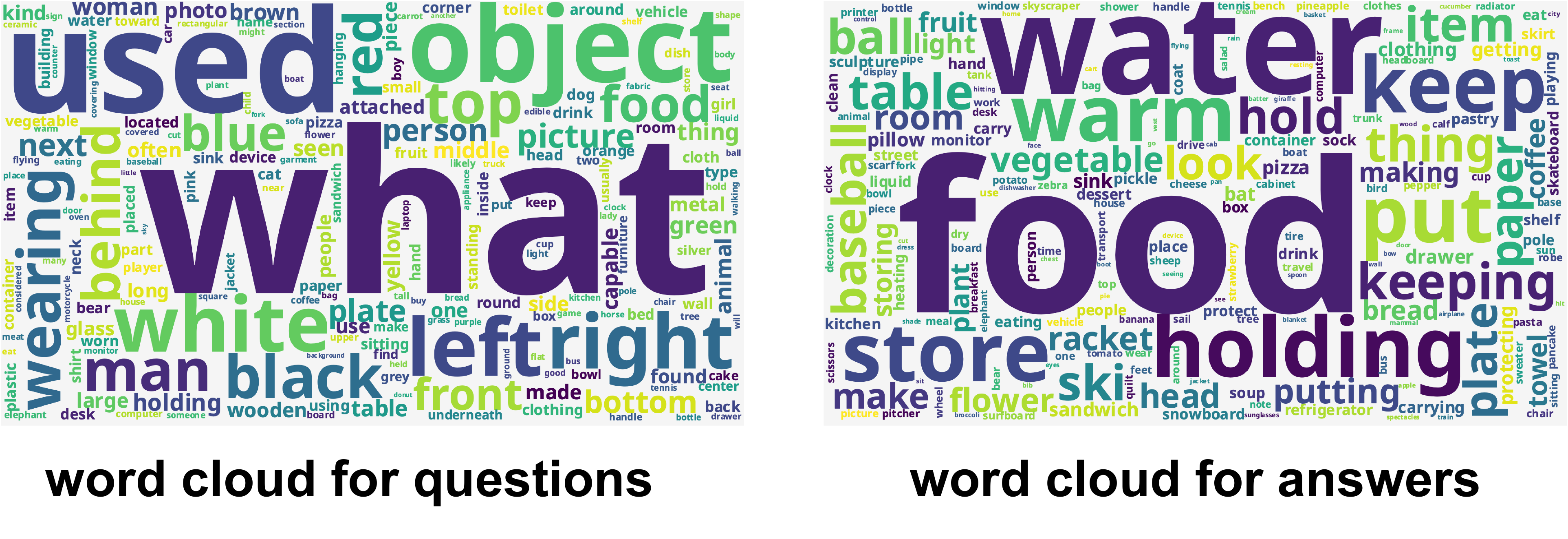}
\caption{Word clouds for the questions and answers in the Professional K-VQG dataset.
}
\label{fig:wordcloud}
\end{figure}
\begin{figure}[t]
    \centering
    \includegraphics[width=0.6\linewidth]{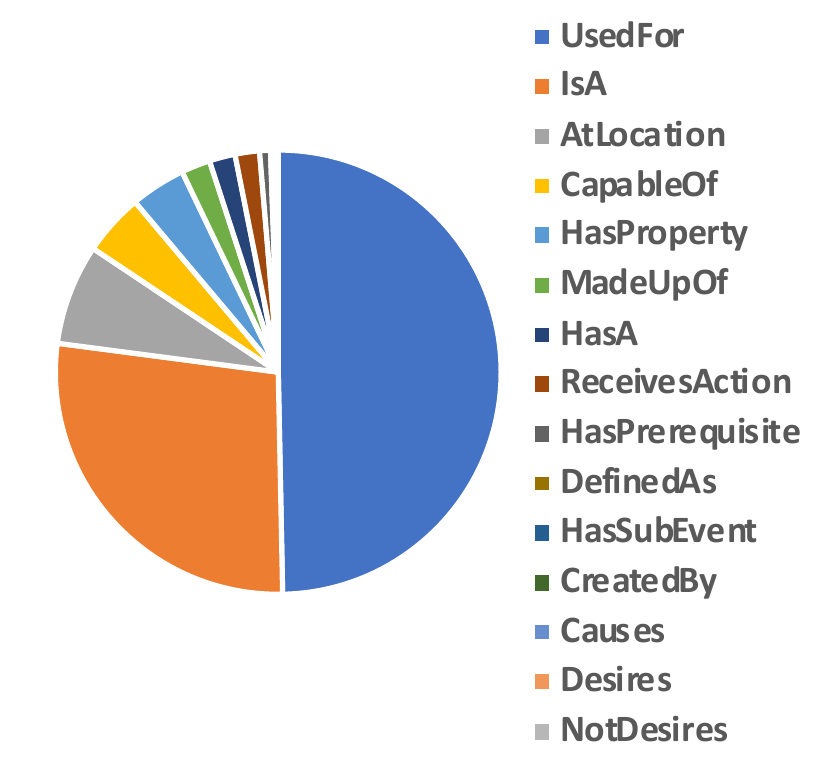}
    \caption{The distribution of relations in the Professional K-VQG dataset.
    }
    \label{fig:relation-count}
\end{figure}

\begin{figure}[t]
    \centering
    \includegraphics[width=1.0\linewidth]{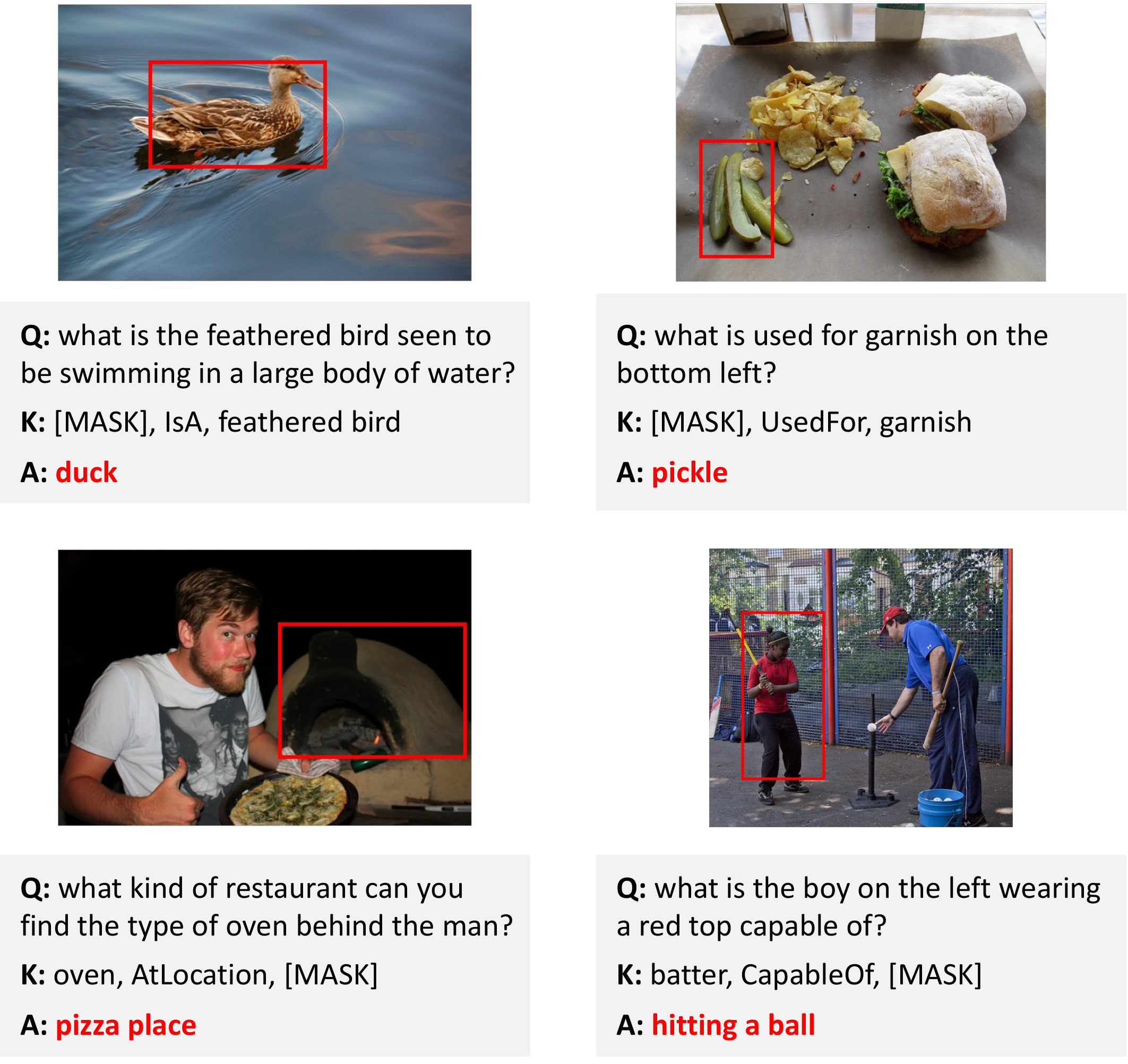}
    \caption{Some examples of the Professional K-VQG dataset.
    }
    \label{fig:pkvqg-pickup}
\end{figure}

\begin{table}[t]
\centering
\scalebox{0.8}{
\begin{tabular}{@{}lccc@{}}
\toprule
 & K-VQG & \begin{tabular}[c]{@{}c@{}}Processional\\ K-VQG\end{tabular} & Merged \\ \midrule
\# questions & 16,098 & 10,431 & 22,212 \\
-- head answers & 11,588 & 5,047 & 13,916 \\
-- tail answers & 4,510 & 5,384 & 8,296 \\
\# images & 13,648 & 9,494 & 9,210 \\ \midrule[0.3pt]
\# unique answers & 2,819 & 3,687 & 4,953 \\
\# unique knowledge & 6,084 & 5,242 & 7,808 \\ \midrule[0.3pt]
\# unique head & 527 & 371 & 533 \\
\# unique tail & 4,922 & 4,257 & 6,199 \\ \bottomrule
\end{tabular}
}
\caption{Detailed statistics of the Professional K-VQG dataset.}
\label{tab:dataset-stats}
\end{table}

To train the QG model, we used the K-VQG dataset~\cite{Uehara2022KVQGKV}, which contains knowledge-aware questions about objects in images, and the newly created \textbf{Professional K-VQG} dataset.
The Professional K-VQG dataset contains knowledge-aware visual questions related to objects annotated by professional annotators.
The source of the images is the Visual Genome~\cite{visualgenome}, and the sources of knowledge are ConceptNet~\cite{conceptnet} and \textsc{Atomic$^{20}_{20}$}~\cite{comet2020}.
We selected 371 object classes from the Visual Genome dataset that were commonly included as heads in the knowledge sources.
In the annotation process, for each candidate object in the image, we obtained the corresponding knowledge triplets from the knowledge sources and asked the annotator to write a knowledge-aware question, with the head or tail of the knowledge triplet as the answer.
As a result, we obtained 10,431 questions for 9,210 images and 5,242 unique knowledge.

We show the word clouds and the relation distribution in the Professional K-VQG dataset in Figure~\ref{fig:wordcloud} and~\ref{fig:relation-count} to visualize several features of the dataset.
We can see that our dataset contains questions on a wide variety of topics about food, clothing, furniture, and so on.
The most and second most frequent relations, ``UsedFor'' and ``IsA'', account for nearly 50\% and 25\% of the total, respectively, which seems a little biased, but this is due to the fact that these relations are also frequent in our knowledge sources.
We show some examples of the Professional K-VQG dataset in Figure~\ref{fig:pkvqg-pickup}.

Because K-VQG and Professional K-VQG share the same annotation format, we merged them to train our QG model.
To use region-based image features obtained from the Faster R-CNN~\cite{butd,Ren2015FasterRT} in the QG module, we excluded samples with target regions that were not detected as objects by the Faster R-CNN (i.e., the IoU between the detected bounding box and the target bounding box was less than 0.5).
We split the merged K-VQG into train, query, and test sets.
The train set was used to train the OC and QG models and the Oracle Answerer.
The query set was used by the OC model to acquire new knowledge, that is, to evaluate the OC model, generate questions to acquire new knowledge using the QG model, and provide answers using the Oracle Answerer.
The test set was used to evaluate the final performance of the OC model using knowledge acquired from the query set.
The train / query / test splits contained 13,334 / 3,693 / 1,414 images and 16,110 / 4,561 / 1,541 questions, respectively.
The number of known knowledge (i.e., knowledge contained in the train set) was 460, and the number of novel knowledge (i.e., knowledge not contained in the train set) was 49.
We show some detailed statistics of the Professional K-VQG dataset in Table~\ref{tab:dataset-stats}.

To train the OC model, we used a dataset consisting of object annotations in the Visual Genome as well as the knowledge triplets obtained from the aforementioned knowledge sources.
The train and validation splits contained 52,321 / 13,080 images and 111,027 / 35,430 objects, respectively, with an average of 52.18 knowledge for each object.
We took care not to contaminate the images and object classes in the train split with other splits in the K-VQG dataset.

\subsection{Training}\label{subsec:training}
We used the same text encoder as CLIP~\cite{clip} and ViT-B/32~\cite{dosovitskiy2020vit} as the visual encoder in the OC model.
All the models were trained using the AdamW optimizer~\cite{adam}, where the learning rate was set to 8e-5 for the OC model and 5e-5 for the QG model.
We trained the OC model for 200 epochs and the QG model for 50 epochs with 8$\times$Tesla A100 GPUs.

\subsection{Baselines}\label{subsec:baselines}
We compared our approach with four baselines: \textbf{CLIP-Ret.}
In this setting, no knowledge acquisition is performed by the QG model, and the performance of the OC model trained with only the training set is evaluated.
\textbf{All Exp. / All Conf.}
In these settings, the question generation policy is fixed to ``exploration'' and ``confirmation,'' respectively.
\textbf{Random Policy.}
The question generation policy is chosen randomly.
We tested this method three times with different random seeds.

\subsection{Results}\label{subsec:results}

\setlength{\tabcolsep}{1.7mm}
\begin{table*}[]
\centering
\scalebox{0.8}{
\begin{tabular}{@{}lcccccc|cc@{}}
\toprule
 & \multicolumn{2}{c}{Overall acc.} & \multicolumn{2}{c}{Known acc.} & \multicolumn{2}{c|}{Novel acc.} &  &  \\
 & zero-shot & fine-tune & zero-shot & fine-tune & zero-shot & fine-tune & \#valid Q. & \#knowledge \\ \midrule
Baseline (CLIP-Ret.) & 38.9 & - & 59.9 & - & 0.0 & - & - & - \\
All Conf. & \textbf{56.2} & 59.7 & \textbf{65.8} & 67.5 & \textbf{38.5} & 45.3 & 9756 & 8272 \\
All Exp. & 41.5 & 45.5 & 57.1 & 54.7 & 12.5 & 28.4 & 1057 & 8431 \\
Random Policy & 48.9 $\pm$ 3.8 & 53.2 $\pm$ 0.39 & 60.6 $\pm$ 2.4 & 60.8 $\pm$ 0.71 & 27.4 $\pm$ 7.1 & 39.2 $\pm$ 0.56 & 2788 $\pm$ 116 & 8521 $\pm$ 169 \\
Ours & 56.1 & \textbf{62.0} & \textbf{65.8} & \textbf{67.5} & 38.3 & \textbf{51.9} & \textbf{9864} & 8354 \\ \bottomrule
\end{tabular}
}
        \caption{
    Results of the OC model.
The top row shows the baseline results without knowledge acquisition through question generation, and the subsequent rows show the results using different question generation policies.
}
    \label{tab:main-results}
\end{table*}

The main results are summarized in Table~\ref{tab:main-results}.
We report results separately for the ``zero-shot'' and ``fine-tune'' settings, the former in which the obtained knowledge is simply added to the OC and tested without additional training, and the latter in which the OC is additionally trained with the obtained knowledge.
Known acc. is the accuracy for data where the correct answers are the objects already contained in the training data (i.e., known objects), and Novel acc. is the accuracy for data where the correct answers are the objects not contained in the training data (i.e., novel objects).

Overall, the proposed methods outperformed the baseline CLIP-Ret on all metrics.
In the zero-shot setting, All Conf. performed best, while Ours performed best in the fine-tune setting.
This may be due to the nature of the confirmation mode.
In the confirmation mode, the question target of the QG model is the partial knowledge (i.e., head masked knowledge) that scored highest in the prediction of OC.
Therefore, when no additional training is performed, the confirmation mode, which allows the acquisition of knowledge that is known to be "useful" in advance, tends to perform better.
However, in the confirmation mode, it is not possible to obtain knowledge with a combination of relation and tail that does not exist in the training data, and the variety of knowledge to be obtained is limited.
Thus, the number of knowledge (\#knowledge) obtained in the confirmation mode was less than in the other modes, which may have resulted in inferior performance after the additional training.

On the other hand, in settings All Exp. and Random Policy, the number of knowledge acquired was larger than in Ours, but the number of valid questions (\#valid Q.) was smaller.
Unlike the confirmation mode, the exploration mode is a mode that is more likely to yield completely new knowledge.
However, in the exploration mode, less information is input to the question generation model and the question generation performance is degraded.
For these reasons, the number of valid novel data available for training has decreased, and the performance of the OC model cannot be adequately improved.

\begin{figure*}[t]
    \centering
    \includegraphics[width=1.0\linewidth]{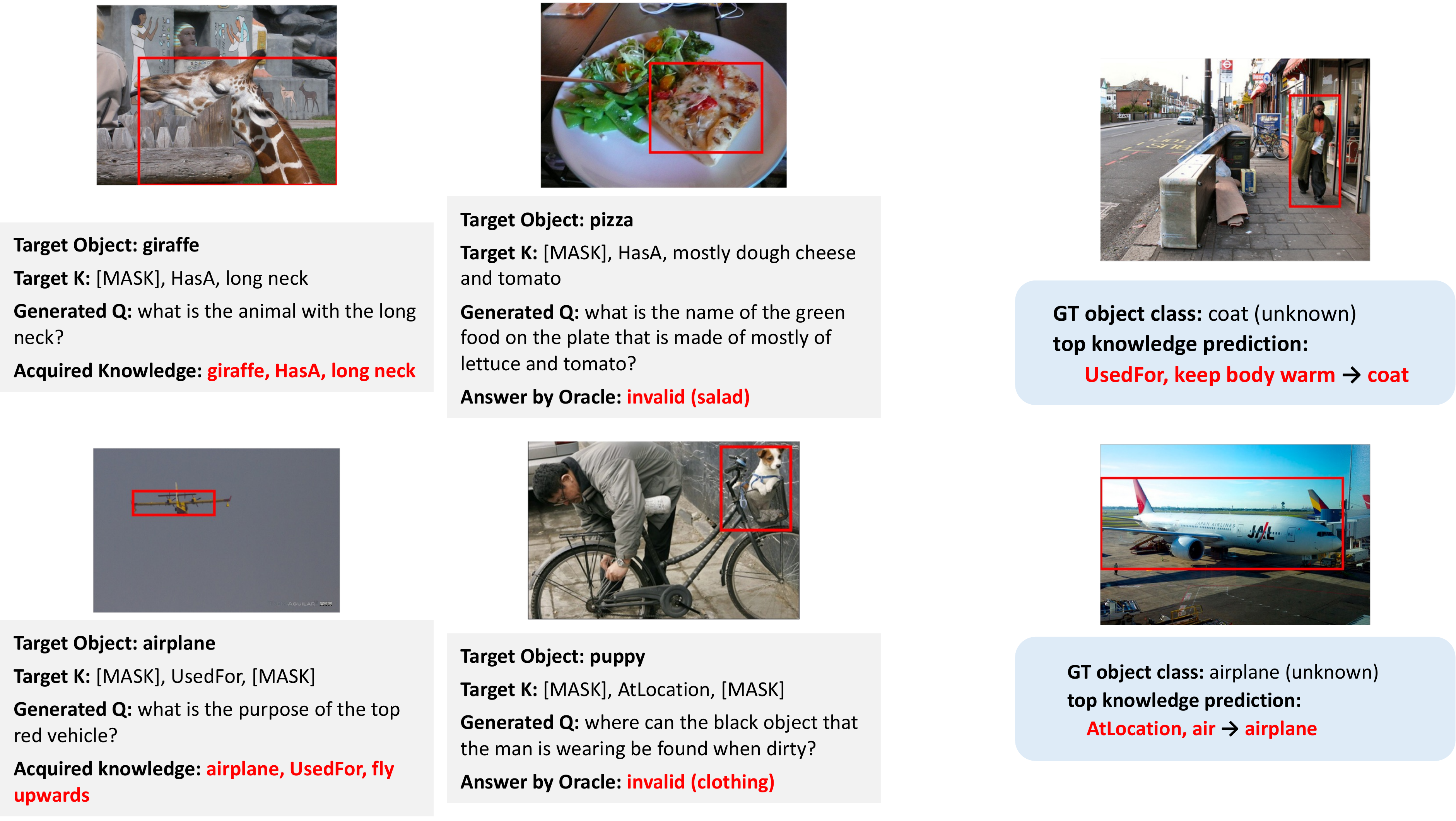}
    \caption{
    Examples of generated questions (left) and the novel object recognition with newly acquired knowledge (right).
       The left column of the question generation results shows examples of successfully acquiring new knowledge.
       In the object recognition results, both ``coat'' and ``airplane'' are novel objects, but the model succeeds to recognize them by using the newly acquired knowledge.
    }
    \label{fig:results}
\end{figure*}

\subsection{Qualitative Results}\label{subsec:qualitative-results}
In Figure~\ref{fig:results}, we show several examples of the generated questions and novel object recognition using the newly acquired knowledge.
As shown in the left column of the question generation results, the QG model generated questions that could successfully acquire new knowledge.
However, in some cases, the QG model failed to generate valid questions.
This is mainly because VQG targeting knowledge is still a challenging task, and the model generates questions that are not relevant to the content of the image or the targeted knowledge.
We leave further improvement of knowledge-targeted question generation as a future issue.

\section{Conclusion}\label{sec:conclusion}
In this study, we presented a novel method for knowledge acquisition through VQG in object classification tasks.
By using a method that appropriately determines the target of the question, we achieved novel object recognition based on knowledge, while reducing the cost of retraining.

A future challenge would be to study better question-targeting policies.
Our work is related to active learning, in which the main topics are to be labeled by a human.
However, when active learning is applied to complicated tasks such as those with multimodal inputs, it is known that the performance gains are smaller than those for random baselines~\cite{Karamcheti2021MindYO}.
We believe that an interesting issue to be addressed in the future is how to apply active learning methods to our proposed task of knowledge acquisition through question generation.

\section*{Acknowledgments}
This work was partially supported by JST AIP Acceleration Research JPMJCR20U3, Moonshot R\&D Grant Number JPMJPS2011, CREST Grant Number JPMJCR2015, JSPS KAKENHI Grant Number JP19H01115, JP20H05556 and Basic Research Grant (Super AI) of Institute for AI and Beyond of the University of Tokyo.
We would like to thank Yusuke Mori for their helpful discussions.

\bibliography{extracted}
\bibliographystyle{acl_natbib}

\end{document}